%% file: nlp4convai21.tex
\pdfoutput=1
\documentclass[11pt]{article}

\usepackage{emnlp2021}

\usepackage{times}
\usepackage{latexsym}

\usepackage[T1]{fontenc}
\usepackage[utf8]{inputenc}

\usepackage{microtype}

\usepackage{amsmath,graphicx}
\usepackage[utf8]{inputenc}
\usepackage{xcolor}
\usepackage{booktabs}
\usepackage{multirow}
\usepackage[inline]{enumitem}
\usepackage[justification=centering]{caption}
\usepackage{bbm}
\usepackage{dblfloatfix}
\usepackage{url}

\title{Using Pause Information for More Accurate Entity Recognition}

\author{Sahas Dendukuri$^1$\thanks{\ \ work done as an intern at Apple},  Pooja Chitkara\thanks{\ \  work done while at Apple}, Joel Ruben Antony Moniz$^2$, Xiao Yang\footnote[2]{},  \\ \textbf{Manos Tsagkias}$^2$, \textbf{Stephen Pulman}$^2$ \\
$^1$University of Michigan, $^2$Apple \\
sahasd@umich.edu, chitkarapooja@yahoo.com, joelrubenantony\_moniz@apple.com,\\ graceyx.scut@gmail.com,
  \{etsagkias, spulman\}@apple.com }

\begin{document}

\maketitle

\begin{abstract}
 Entity tags in human-machine dialog are integral to natural language understanding (NLU) tasks in conversational assistants. However, current systems struggle to accurately parse spoken queries with the typical use of text input alone, and often fail to understand the user intent. Previous work in linguistics has identified a cross-language tendency for longer speech pauses surrounding nouns as compared to verbs. We demonstrate that the linguistic observation on pauses can be used to improve accuracy in machine-learnt language understanding tasks. Analysis of pauses in French and English utterances from a commercial voice
 assistant shows the statistically significant difference in pause duration around multi-token entity span boundaries compared to within entity spans. Additionally, in contrast to text-based NLU, we apply pause duration to enrich contextual embeddings to improve shallow parsing of entities. Results show that our proposed novel embeddings improve the relative error rate by up to 8\% consistently across three domains for French, without any added annotation or alignment costs to the parser.
 
\end{abstract}
\noindent\textbf{Index Terms}: Natural Language Understanding, BERT, Shallow Parsing, Entity Recognition, Contextual Embeddings

\input{01-introduction}

\input{02-pause-duration}

\input{03-modeling}

\input{04-exp-design}

\input{05-results}

\input{06-conclusions}

\input{07-acknowledgements}

\bibliography{nlp4convai21}
\bibliographystyle{acl_natbib}

\end{document}

%% file: 01-introduction.tex
\section{Introduction and Related Work}
\label{section:introduction}

\begin{table*}
\caption{Statistics for the dataset EngFrPauseData.} 
\centering
\setlength{\tabcolsep}{15pt}
\begin{tabular}{ c c c }
\toprule
\textbf{Measure}                    &\textbf{French}    &\textbf{English}   \\
Number of utterances                & 17,799            & 22,388    \\
Average \# tokens in an utterance   & 3.67              & 4.71      \\
\% of tokens that are entities      & 24\%              & 24.52\%  \\
Average pause duration per token    & 51.60ms           & 33.64ms   \\
\bottomrule
\end{tabular}
\label{tab:overview}
\end{table*}

Voice assistants are increasingly becoming a part of our daily lives as they empower us to complete diverse and complex tasks. Today, voice assistants are able to make phone calls, take notes, set alarms, narrate bed time stories, or even rap hip-hop. Given a user's request, the assistant transcribes the input audio to text, and a natural language understanding (NLU) system classifies and routes the transcribed text to  components that handle the request (e.g., phone call, notes, alarm). 

Typical conversational NLU systems rely on text, the output of an automatic speech recognition (ASR) system \cite{Muralidharan2019LeveragingUE}. These systems have proven successful, but relying wholly on text means that some analyses may be inaccurate \cite{nygaard2009semantics}, since acoustic signals may be needed to arrive at the right intent. For example, in the utterance ``play thank you next'', the pause duration between ``you'' and ``next'' can help determine if the request is to play the specific song ``thank you next'' or to play the song ``thank you'' but next in the queue. This disambiguating function of pause information is well known, but the potential role of pause information in signalling different types of constituents, as described in \citet{Seifart5720},  has not, to our knowledge, been investigated in this setting.

Our goal is two-fold. Our first objective is to analyze the findings of \citet{Seifart5720} in the context of spoken conversations typical of those processed by voice assistants. Specifically, we show that differences in pause length around named entity spans is a statistically reliable processing cue.
Our second goal is to demonstrate that the intuitive role of pause duration in aiding language understanding can be used in practice. Specifically, we present novel pause-grounded contextual embeddings that consistently outperform text-based representations in a language understanding task across multiple domains.

A standard approach to a tighter integration between acoustic information and NLU has been to use end-to-end models that can consume acoustic waveforms directly and learn NLU labels \cite{haghani2018audio}. However, this approach can be difficult to adapt to traditional text-based tasks since audio inputs need to be labeled with task-specific information, introducing additional annotation costs.  

An alternative approach is suggested by the recent and growing body of research which encodes text along with other modalities, such as audio and video, in a single embedding, known as multi-modal grounded embeddings~\cite{settle-2019, jeon-2019, rahman-etal-2020-integrating, lan-2020, chuang2019speechbert, chung2020semisupervised}. These models have proven successful in tasks like speech recognition~\cite{settle-2019} and sentiment analysis~\cite{rahman-etal-2020-integrating}. 
However, they require the data from different modalities to be aligned in order to produce the embeddings, again increasing the cost of generating training data.

We work within this paradigm of multi-modal grounded embeddings using pause markers in spoken utterances for the task of \emph{shallow parsing}~\cite{ladhak-2016,vijayakumar-etal-2017-sound, rahman-etal-2020-integrating, liu-2020, lan-2020}. We propose a novel self-supervised architecture adding a pause prediction task to traditional text-based BERT-style language models~\cite{devlin-etal-2019-bert}. To the best of our knowledge, this is the first study on grounding textual embeddings in spoken pause signals.

We pre-train our embeddings using token-level pauses from the ASR system, and consume the representations in the downstream parsing task without fine-tuning. We do not input pause duration in our downstream task and, therefore, do not require data from different modalities to be aligned. This sets our approach apart from end-to-end approaches that use speech signals as input during inference, requiring additional task-specific annotation (such as  \citet{chi2020audio, liu-2020,ling2020bertphone}).

%% file: 02-pause-duration.tex
\section{Analysis of Pause Duration}
\label{section:pauseduration}

 \begin{figure*}[tb]
 \begin{minipage}[b]{.48\linewidth}
    \centering
    \centerline{\includegraphics[width=8.0cm]{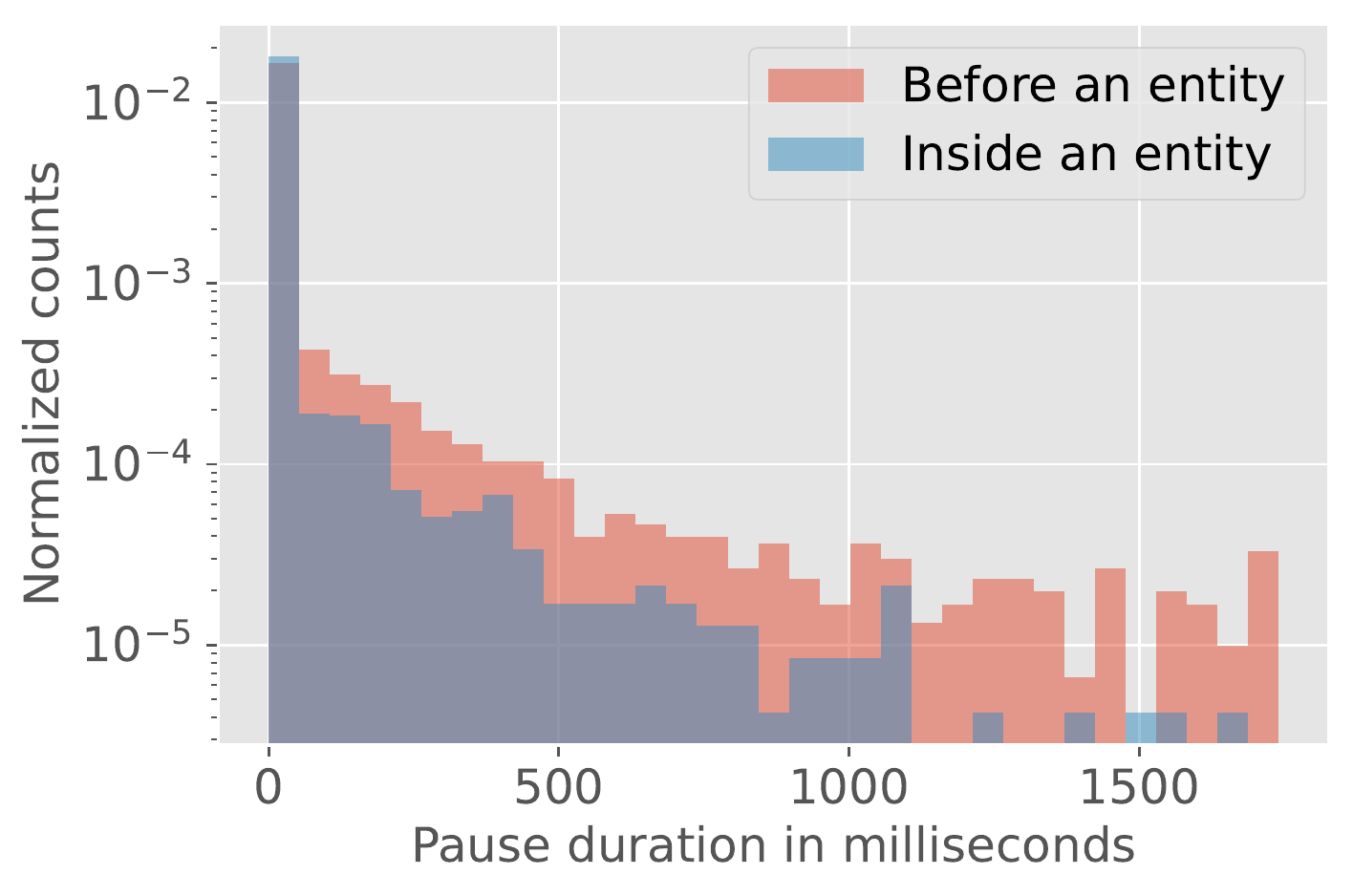}}

    \centerline{(a)}\medskip
 \end{minipage}
 \begin{minipage}[b]{0.48\linewidth}
    \centering
    \centerline{\includegraphics[width=8.0cm]{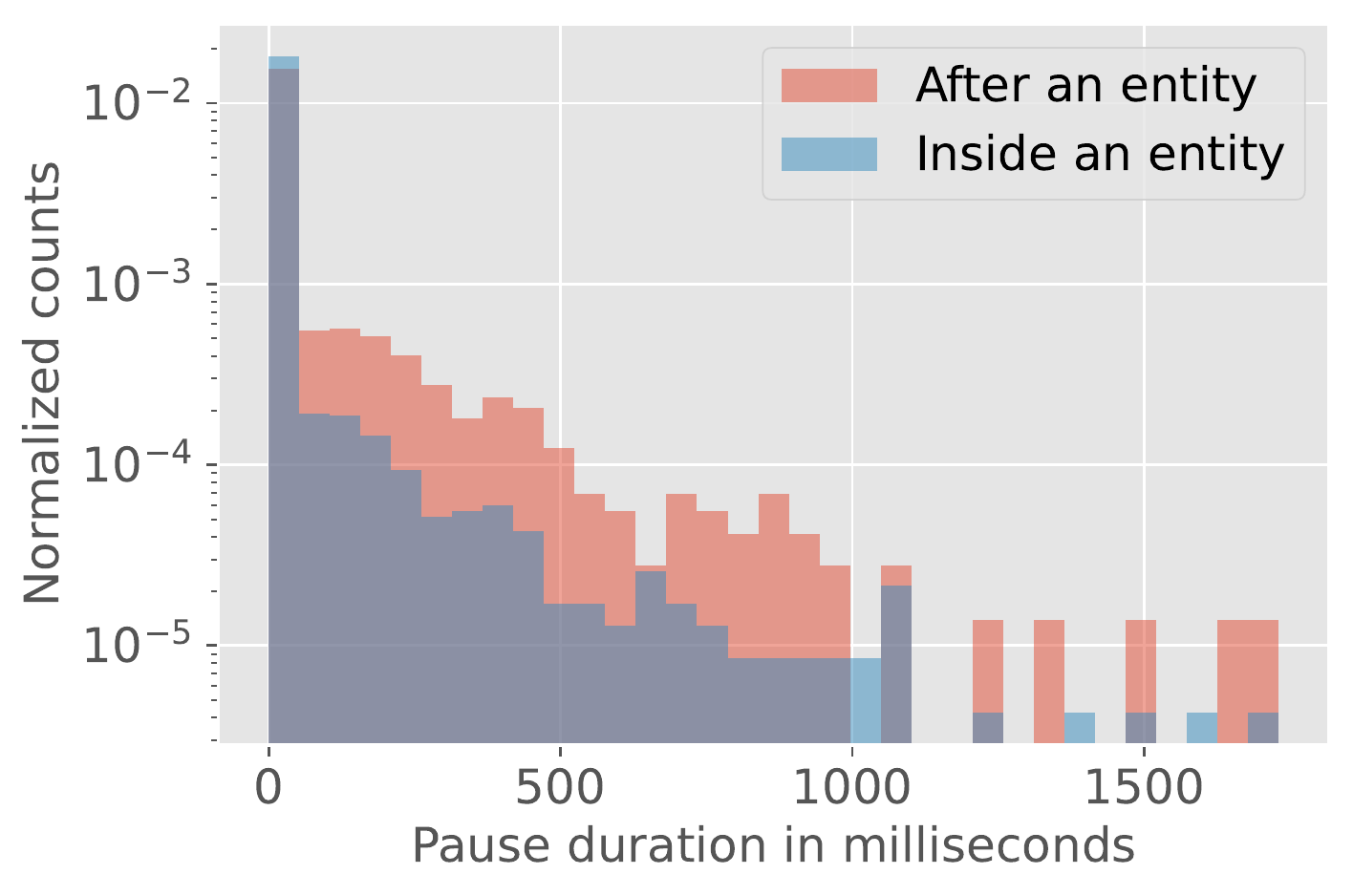}}
    \centerline{(b)}\medskip
 \end{minipage}
 
  \caption{Histograms showing the normalized frequency distribution of the pause duration for French on the EngFrPause Data described in Table \ref{tab:overview} across three domains: sports, movies, and music. x-axis marks pause duration in milliseconds (up to 3 standard deviations from the mean) for before (O-B) and inside (B-I and I-I) an entity (left), and after (B-O and I-O) and inside (B-I and I-I) an entity (right). y-axis shows log-normalized counts. Note that the deep purple-grey color is indicative of the red and blue histograms overlapping.}
 
 \label{fig:distribution-pauses}
\end{figure*}

 \begin{figure*}[t]
 \begin{minipage}[b]{.48\linewidth}
    \centering
    \centerline{\includegraphics[width=8.0cm]{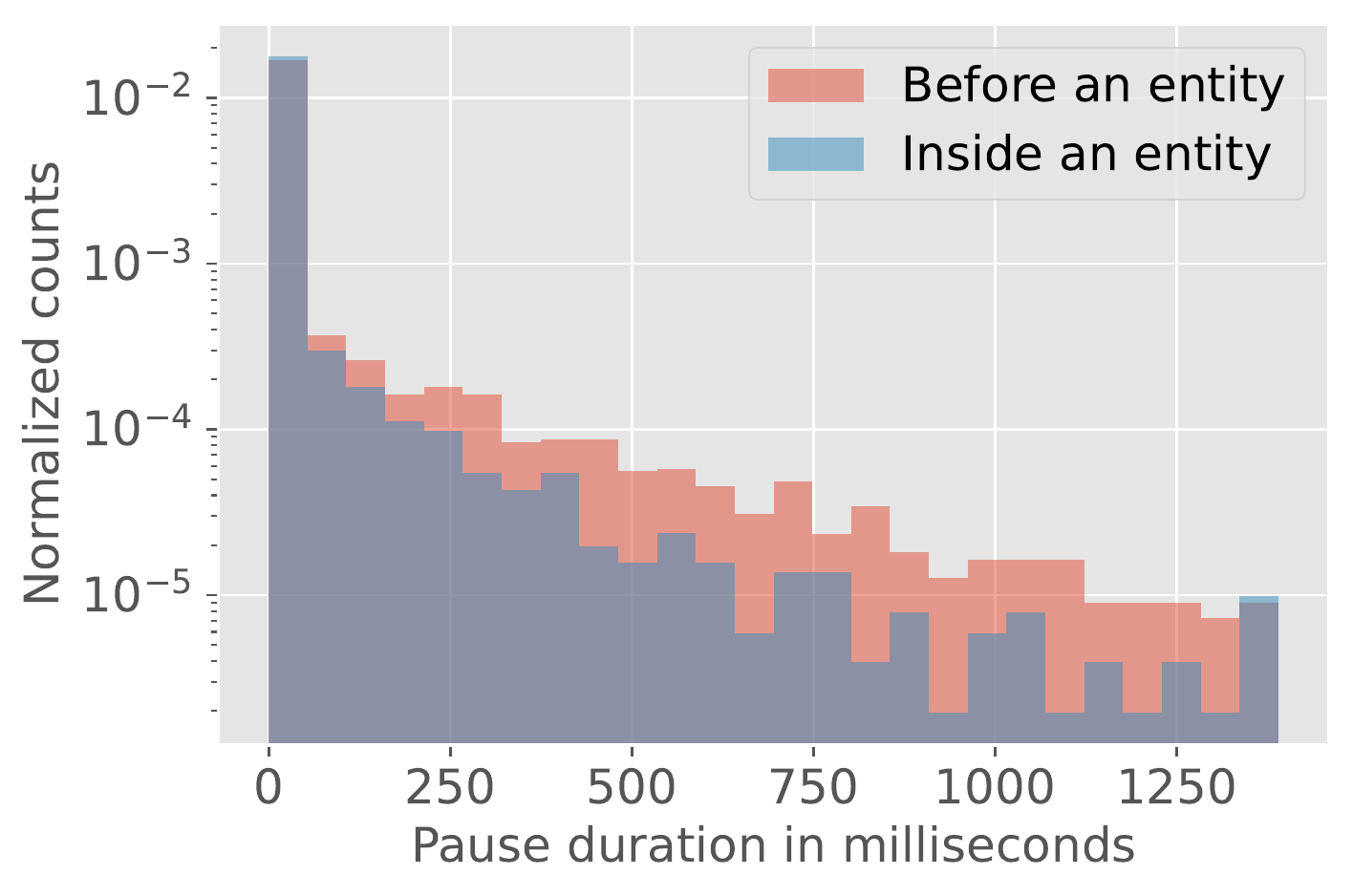}}

    \centerline{(a)}\medskip
 \end{minipage}
 \begin{minipage}[b]{0.48\linewidth}
    \centering
    \centerline{\includegraphics[width=8.0cm]{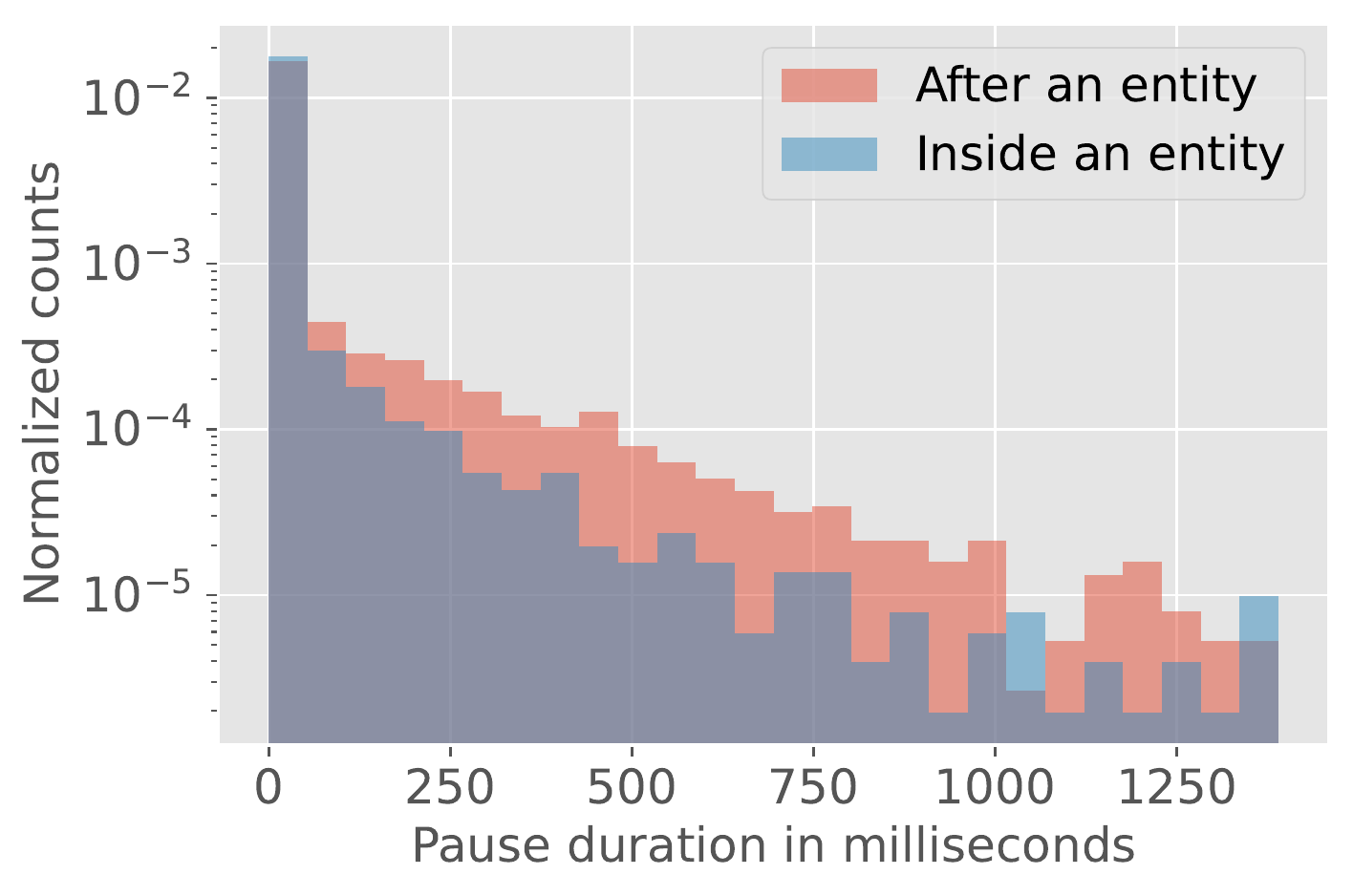}}
    \centerline{(b)}\medskip
 \end{minipage}
   \caption{Histograms showing the normalized frequency distribution of the pause duration for English on the EngFrPause Data described in Table \ref{tab:overview} across three domains: sports, movies, and music. x-axis marks pause duration in milliseconds (up to 3 standard deviations from the mean) for before (O-B) and inside (B-I and I-I) an entity (left), and after (B-O and I-O) and inside (B-I and I-I) an entity (right). y-axis shows log-normalized counts. Note that the deep purple-grey color is indicative of the red and blue histograms overlapping.}
    \label{fig:distribution-pauses-en}
\end{figure*}

In \citet{Seifart5720}, time-aligned spoken language corpora in multiple languages were compared in terms of speech rate and pause duration. The
variable of interest was the difference between nouns and verbs, with the subtlety that nominalisations of verbs still
counted as verbs, and (less frequent) verbalisings of nouns counted as nouns. This means the the distinction was close
to the traditional semantic distinction of `object' and `action' words. Their main findings were that speech rate slows relatively before nouns, and for most languages, pauses are more likely before nouns than before verbs.

We are most interested in the observation about pauses before nouns, as a potential signal for an upcoming named entity sequence. Named entities may not be limited to single nouns, but may be a span of tokens with at least a head noun. We analyze if pauses are more likely and/or longer before (or after) the boundary of an entity span than within the span itself.

We investigate this as follows. We consider a set of utterances from a large internal, anonymous dataset,
collected over several months and domains (referred to throughout as UsageData). Given the sampling methodology, it can be concluded that there are no speaker-dependent effects in our dataset. The utterances have been taken from an upstream ASR
system which provides each utterance hypothesis with pause durations following each token, in milliseconds. In order to
test the findings of \citet{Seifart5720} in the voice assistant setting, we randomly sample UsageData utterances from the sports,
movies, and music domains in  English and French (referred as EngFrPauseData); see Table~\ref{tab:overview} for an overview.

 \begin{figure*}[t]
 \centering
 \begin{minipage}[b]{.4\linewidth}
    \centering
    \centerline{\includegraphics[width=.98\linewidth]{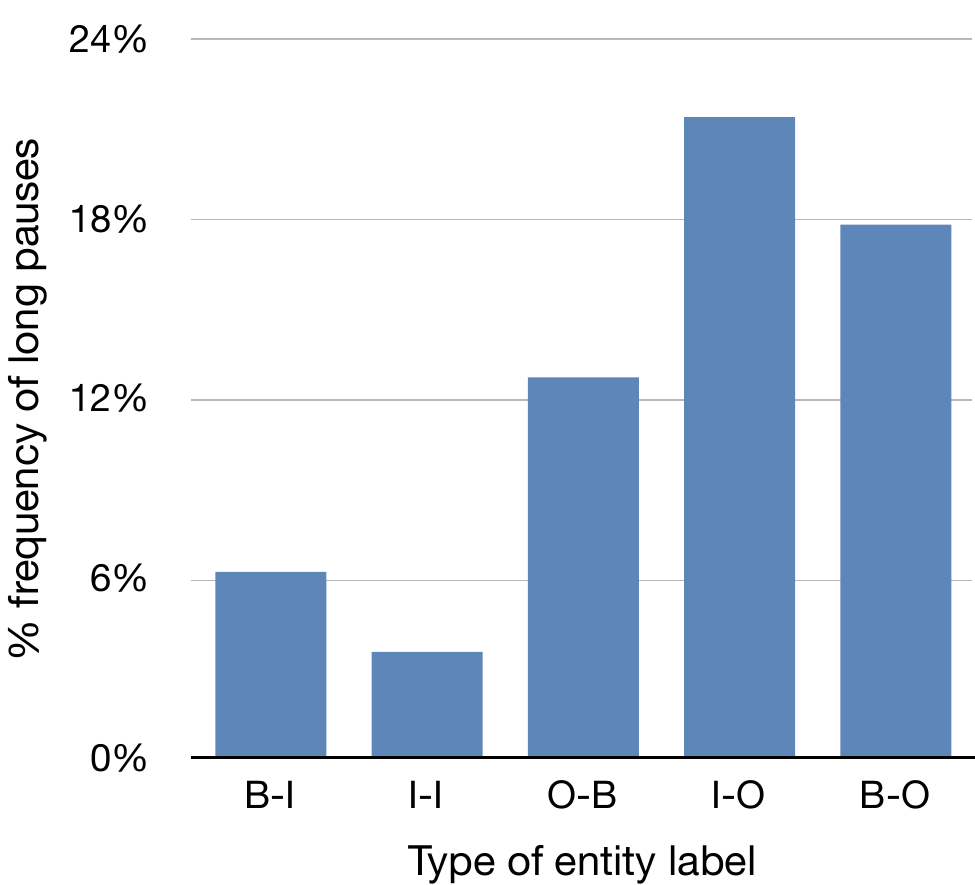}}
    \centerline{(a)}\medskip
 \end{minipage} \qquad
 \begin{minipage}[b]{0.4\linewidth}
    \centering
    \centerline{\includegraphics[width=.98\linewidth]{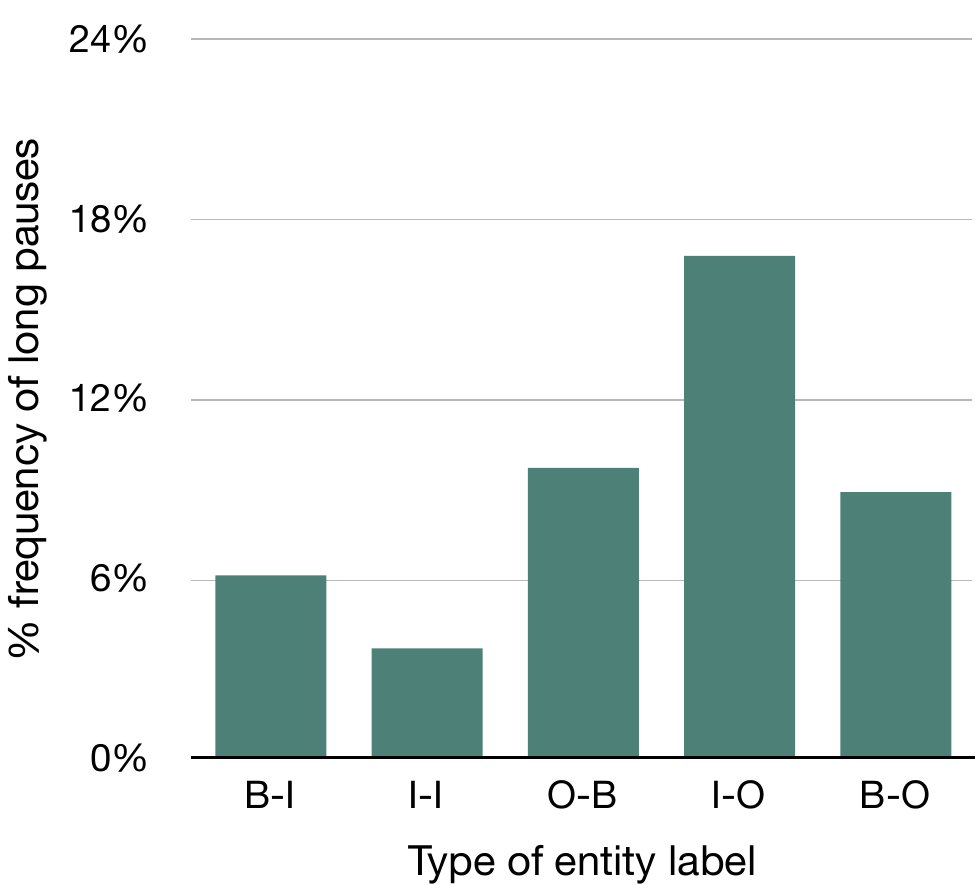}}
    \centerline{(b)}\medskip
 \end{minipage}
   \caption{Histogram of percentage frequency of pause duration >=60ms on the EngFrPause Data (described in Table \ref{tab:overview}) per label type for (a) French and (b) English. x-axis marks entity label type and y-axis shows percentage frequency of long pauses. O-B, I-O, and B-O indicate label types at entity span boundaries (before/after entites), and B-I and I-I indicate label types within entity spans.}
    \label{fig:nonzero}
\end{figure*}

\begin{figure*}[b]
\centering
\begin{minipage}[b]{.7\linewidth}
  \centering
  \centerline{\includegraphics[width=.98\linewidth]{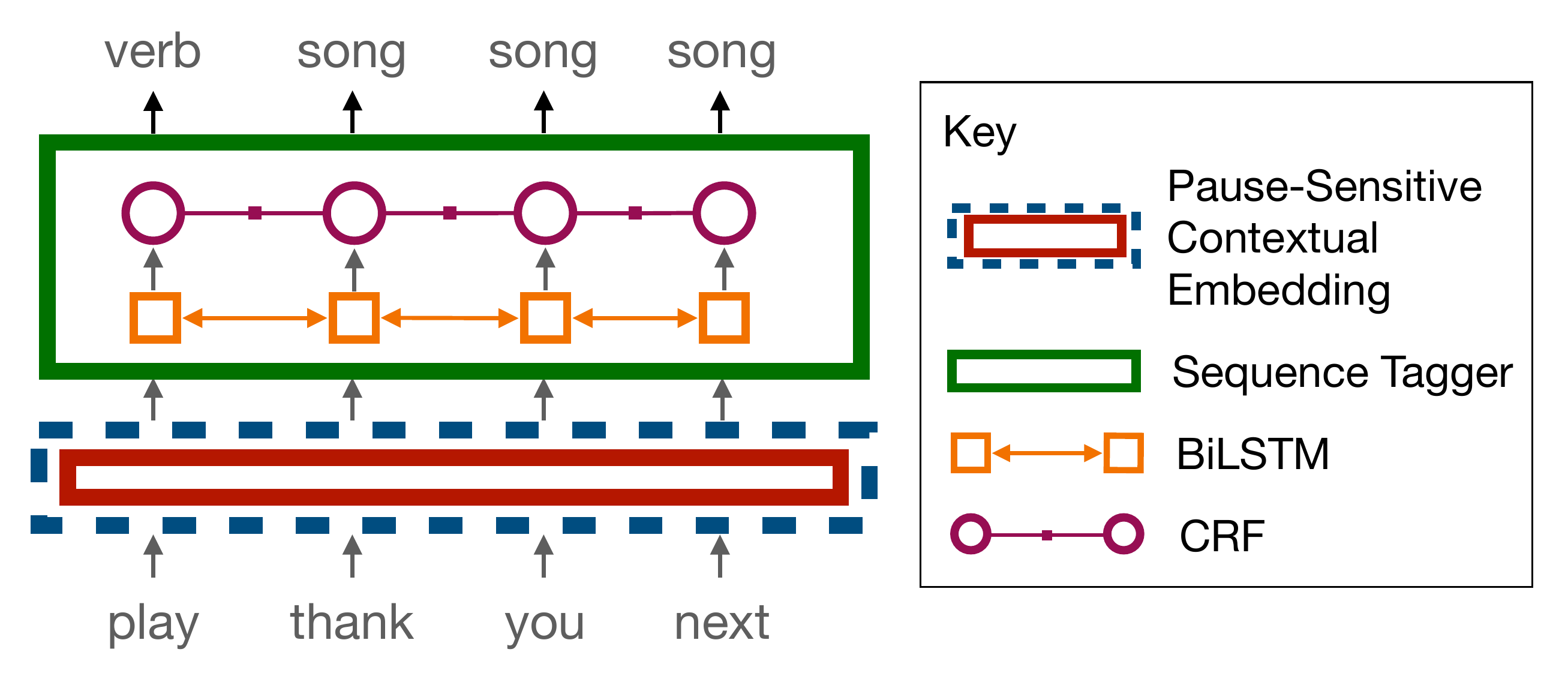}}
\end{minipage}
\caption{Illustration of our shallow parser module.}
\label{fig:shallow}
\end{figure*}

We manually annotate EngFrPauseData with BIO-style tags \cite{ramshaw-marcus-1995-text,sang2003introduction}; O for non-entity, B for beginning of an entity, I for inside an entity. Furthermore, if a token is tagged with a B or I, it contains an additional entity tag (e.g., Artist). We combine the pause information from ASR with the named entity tags to derive pauses before, during, and after an entity. We aggregate pause and token duration statistics across all utterances for all domains in a language and fit distributions for the following pairs of settings:
O-B,
B-I,
I-I,
B-O,
I-O.
For example, an O-B pair marks the pause duration between a token tagged O and the token following it tagged B, while an I-I pair denotes the pause between a token tagged I and the token following it also tagged I.

\begin{figure*}[tb]
\centering
\begin{minipage}[b]{.7\linewidth}
  \centering
  \centerline{\includegraphics[width=.98\linewidth]{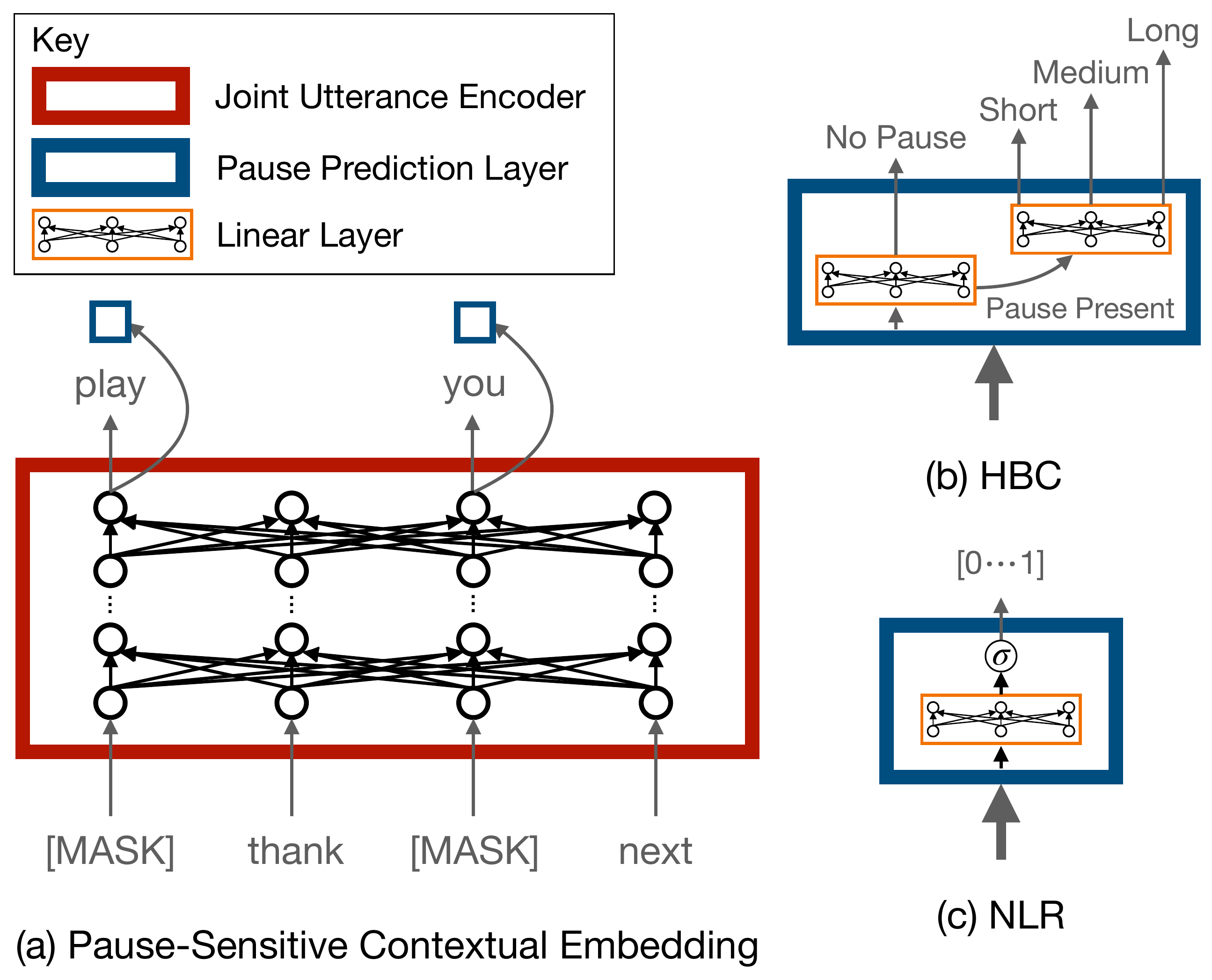}}
\end{minipage}
\caption{Pause-Sensitive Contextual Embedding. (a) Joint Utterance Encoder, in which the prediction layer can be (b) HBC, (c) NLR, or absent (baseline).}
\label{fig:model-diagram}
\end{figure*}

 To analyze whether the study in \citet{Seifart5720} can be extended to spoken entity sequences, we ask whether the pause duration before,
 within, and after an entity is different in a statistically significant way. 
To do this, we first log normalize the frequency of pauses before, inside, and after an entity to account for differences in their raw counts. In Fig.~\ref{fig:distribution-pauses} and Fig.~\ref{fig:distribution-pauses-en} we plot and compare histograms showing the normalized frequency distributions of pauses
 before an entity (O-B) and inside an entity (B-I and I-I), and inside an entity and after an entity (B-O and
 I-O) for
 French and English respectively. We observe that pauses before uttering and after uttering an entity are longer than pauses within the entity
 itself.
 
 For French, the mean duration for pause within an entity is 18.17ms, for pause before an entity is 55.04ms, and
 for pause after an entity is 63.86ms. The differences in mean pause duration for both within and before an entity, and
 for within and after an entity are found statistically significant using a two sample statistical significance t-test;
 p-values are 4.59e-34 and 3.97e-20, respectively. 
 
 For English, the mean duration for pause within an entity is 15.21ms,
 for pause before an entity is 33.99ms, and for pause after an entity is 38.08ms.  The differences in pause duration are
 again found to be statistically significant for both before and within an entity (p-value: 7.31e-33), and within and after an
 entity (p-value: 6.81e-36). These findings are in line with those from \citet{Seifart5720} and suggests that acoustic
 pauses are indeed a useful and complementary signal to text-based features.

 Likewise, in Figure~\ref{fig:nonzero}, we show the percentage frequency of pauses greater than or equal to 60ms for different types of entity labels. Here, too, we observe that such pauses are more frequent after or before entities (label types O-B, I-O, B-O) as compared to within entities (label types B-I, I-I), for both French (Figure~\ref{fig:nonzero}a) and English (Figure~\ref{fig:nonzero}b).

 Note that the pause length difference within and outside an entity span tends to be greater in French
 than in English. Similarly, the difference in frequency of longer pauses at entity boundaries versus within entity spans is greater in French than in English, as can be visualized in Figure~\ref{fig:nonzero}. We therefore carried out our further experiments on the French data (hereafter, FrPauseData), both for reasons of time, and  on
 the grounds that any effects were more
 likely to show up there. \footnote{Regrettably, we could not find a public domain dataset with analogous properties so that our
 findings could be replicated by others, and so we have have presented results on this proprietary data and shared the statistics of our dataset.}

%% file: 03-modeling.tex
\section{Shallow Parsing}

Our primary downstream setting that consumes contextual embeddings (described in Section \ref{section:modeling}) comprises of shallow parsing, as a sequence labelling task. It takes text output from an ASR system as the input, and
produces an output label for each input token. The labels aim to recognize the intent of the spoken utterance, typically
picking out a verb and the phrases that could constitute arguments of the verb. For example, ``play thank you next"
would be parsed as ``play/Verb thank/Song you/Song next/Song".\footnote{``thank you next" is the ASR recognition result
  for the song ``Thank u, Next".} 
  
In all experiments, we first pass the text output obtained from ASR into either a pre-trained baseline or pause-sensitive contextual embedding (refer Section \ref{section:modeling}). We pass the so-obtained contextual embedding representations into a relatively simple shallow parsing model. The shallow parser is a sequence tagger model that feeds the contextual embedding corresponding to each text token into a single-layered BiLSTM with a CRF on top, obtaining a predicted
label for each token as the output (refer Fig.~\ref{fig:shallow}). During training, our shallow parser is optimized by minimizing the Negative Log Likelihood (NLL) over the data. Our shallow parser is identical to that of  \citet{muralidharan2021noise}, except that we feed the output of the BiLSTM into a CRF layer, which we empirically observe boosts performance in all cases.

\section{Pre-training Embeddings with Pause Information }

\label{section:modeling}

The baseline version of the shallow parser uses a text-only embedding. This embedding is a BERT-style text-based
language model~\cite{devlin-etal-2019-bert} trained without the next sentence prediction auxiliary task of the
original BERT architecture.

To investigate the effect of pauses, we compare the baseline architecture with an identical BERT-based model extended with an additional spoken pause prediction task, as shown in Fig.~\ref{fig:model-diagram}. We are given, for each utterance, the pauses following each token in milliseconds, and we learn a model jointly optimizing a text-based language modeling task and a token-level pause duration prediction task. We add additional linear projections over intermediary representations from the encoder to predict pause durations following a token. We refer to the encoder as the ``Joint Utterance Encoder'', and to the linear projection as the ``Pause Prediction Layer''. We use negative log-likelihood (NLL) for categorical loss and mean squared error (MSE) for regression loss optimizations. We backpropagate the sum of the losses from both the text-based and the speech-based pause prediction tasks into the Joint Utterance Encoder. 

We consider two variants of the prediction task:\\
{\bf Hierarchical Bin Classification [HBC]}: The task is to perform hierarchical classification where the first linear projection predicts the boolean presence/absence of a pause and the second linear projection maps non-zero pauses (in the case of presence of a pause) into one of three labels: short, medium, and long, indicative of the pause durations. \\
{\bf Normalized Linear Regression [NLR]}: The task is to linearly predict the length of pause (normalized into a range from zero to one) following a token.

More concretely, in a given utterance, let ${T_b}$ be the set of tokens used to train the BERT-style model using a Masked Language Model (MLM) objective and $l^b_i$ be the softmaxed LM prediction for the probability of the $i^{th}$ token $t_i$. 
Further, let ${T_s}$ be the set of tokens on which the pause prediction tasks are to be trained, and for the $t_i^{th}$ token, let $l^c_i$ be the predicted probability of the true coarse-grained (presence/absence) label by HBC, let $l^f_i$ be the predicted probability of the true fine-grained short (S), medium (M), or long (L) label by HBC, let $l^r_i$ be the predicted pause by the NLR module, let $\mathbbm{1}_{g^c_i}$ indicate the presence/absence of a pause, and let $g^r_i$ be the true normalized pause duration. 
Then the baseline BERT-style encoder is trained by minimizing the loss 

\begin{equation}L_{\texttt{BERT}} = -\sum\limits_{t_i \in T_b} \log l^b_i.\end{equation}

\noindent The loss for the HBC task is 

\begin{equation}L_{\texttt{HBC}} = -\sum\limits_{t_i \in T_s} \left( \log l^c_i + \mathbbm{1}_{g^c_i} \log l^f_i \right). \end{equation}

\noindent The loss for the NLR task is 

\begin{equation}L_{\texttt{NLR}} = \sum\limits_{t_i \in T_s} \left(g^r_i - l^r_i\right)^2. \end{equation}

\noindent 
The joint loss is $L_{\texttt{BERT}} + \lambda L_{\texttt{HBC}}$ or $L_{\texttt{BERT}} + \lambda L_{\texttt{NLR}}$ respectively, where $\lambda$ is a weighting factor set to 1 in our experiments.

%% file: 04-exp-design.tex
\section{Experimental Design}
\label{section:exp-design}

    \begin{table*}
    \caption{Relative percent change in error rates for Shallow Parser with pause-sensitive embeddings. Boldface marks best performance for domain and metric.}\label{tab:results}
    \centering
    \setlength{\tabcolsep}{15pt}
    \begin{tabular}{llrrr}
    \toprule
    \textbf{Model} & \textbf{Domain} & \textbf{EER} & \textbf{TER} & \textbf{UER} \\
    \midrule
    \multirow{3}{0em}{HBC} & Music& +0.47\%& +0.22\%& -2.34\% \\
    & Movies& -2.83\% & -2.78\%& -0.57\% \\
    & Sports& +0.70\%& +0.71\%& -1.12\% \\
    \midrule
    \multirow{3}{0em}{NLR} & Music& \textbf{-2.94\%}& \textbf{-3.32\%}& \textbf{-4.10\%} \\
    & Movies& \textbf{-8.32\%}& \textbf{-8.51\%}& \textbf{-3.99\%} \\
    & Sports& \textbf{-2.63\%}& \textbf{-2.67\%}& \textbf{-3.22\%} \\
    \bottomrule
    \end{tabular}
    \end{table*}

In our experiments, we compare the parser with the baseline version of text-only embeddings against a version using
embeddings enhanced with pause information (HBC and NLR). This is the only difference between the two systems: the
speech pause signals used to pre-train the embeddings are not used directly in the parser. Neither the baseline version nor the pause-sensitive embeddings are fine-tuned in the parser.

For pre-training the BERT-style language models, we train over a random sample  from the French section of UsageData without entity annotations (`FrTrainData'). FrTrainData contains about 50 million utterances with token-level ASR feature tuples consisting of recognized text and post-token pause duration (in milliseconds) for each input point.

To generate label boundaries for the HBC task, we sort pause durations in FrPauseData in ascending order and divide the resulting list of pauses into three parts with an equal number of data points. These serve as the label boundaries for the S, M, and L length labels for HBC. Empirically, pause durations less than 60ms are labeled as S, those between 60ms and 310ms are labeled as M, and those greater than 310ms are labeled as L. During training, we discard any pause greater than 10,000ms as noise from upstream ASR. For NLR, we normalize the pauses between 0 and 1 by dividing each duration by 10,000ms (the longest pause duration in the pruned dataset).

For the classification task, we opt for HBC to address the class imbalance problem in our data~\cite{tsagkias-2009} (see Figure~\ref{fig:distribution-pauses}, which shows a large proportion of tokens with zero pause detected). This coarse-to-fine approach helps with learning the presence/absence of pauses (coarse label) over the skewed dataset while also allowing the auxiliary S/M/L prediction (fine label) to be sufficiently fine-grained to prove useful in improving the feature representation.

To train the embeddings, we consider a vocabulary of the 100k most frequent tokens in FrTrainData. To train the pause prediction layers, we predict the labels over a subset of tokens per utterance. We randomly sample 15\% of the tokens per utterance up to a maximum of 3 tokens, including at least one token with a non-empty label (words not contributing to the user intent are given a null label by the shallow parser). We backpropagate the sum of losses from the text-prediction and the pause-prediction tasks to the Joint Utterance Encoder.

We evaluate system performance on shallow parsing with three error rate measures at entity, token, and utterance levels:\\
\noindent
{\bf Entity Error Rate (EER)} measures the proportion of incorrectly labeled entities out of all entities.\\
{\bf Token Error Rate (TER)} measures the proportion of incorrectly labeled entities out of all tokens (both entities and non-entities).\\
{\bf Utterance Error Rate (UER)} measures the proportion of utterances with at least one mislabeled entity. 

\noindent 
We report all metrics as means over 10 runs for each model configuration.

%% file: 05-results.tex
\section{Results}

\label{section:results}

We compare the performance of our pause-sensitive model with that of our baseline text-based BERT architecture in Table~\ref{tab:results}. Our best performing model consistently outperforms the baseline with an improvement in error rates on entity, token, and utterance levels across all three domains under study. 
We note that NLR consistently provides a greater improvement over HBC. This is in line with our expectations since user requests to a voice assistant are typically short utterances with small inter-token pauses (Table \ref{tab:overview}) across domains. Unlike the regression objective of the NLR model which is trained on exact pause durations scaled between 0 and 1, to train over the classification objective of HBC, we discretize these pause durations into four classes losing the granularity of small variations in these recorded pauses.

Looking at performance across domains, we observe that improvement in EER and TER for the  Movies domain is consistently higher than that for Music and Sports. We look at FrPauseData to find that entity names in the Movies domain typically contain longer sequences of tokens than entities for the Music and Sports domains, where bynames and abbreviations are often used. On average, entity spans in the Movies domain are relatively longer than entity spans in Music by 49.64\%, and in Sports by 39.71\% in our dataset. For example, the majority entity span type in the Movies domain is 'movie title' which on average consists of 2.54 tokens and makes 34\% of the domain. This token length is more than double that of the majority label type 'team name' in the Sports domain which makes 60\% of the domain and contains 1.25 tokens on average. We note a similar pattern on observing the top-3 most frequent entity types per domain. For example, the mean token length for the top-3 most frequent labels in the Movies domain is 2.43 tokens compared with only 1.59 tokens in the Music domain.
This relates with our earlier experiment with FrPauseData---we noted a longer mean pause duration of 120.23ms for Movies, compared with 39.06ms for Music, and 29.38ms for Sports. We hypothesize that longer sequences of entity names cause speakers to make longer pauses around these entity spans. 

Finally, since the NLR model improves performance across all domains under study, we emphasize that it can be used in settings where a pre-trained set of embeddings are shared across multiple applications, which is particularly relevant in open-topic voice assistant NLU systems where the domain of a user request is ascertained at run time.

%% file: 06-conclusions.tex
\section{Conclusion}
\label{section:conclusions}
In this paper, we analyzed the statistically significant correlation of pause lengths with syntactic properties in real-world usage data for three entity-rich application domains.  To exploit this property, we presented a novel pause-sensitive contextual embedding model. The model outperforms text-based representations in a language processing task across multiple domains. Our approach does not impose additional task-specific annotation requirements and can be extended to more text-based target tasks. 
To the best of our knowledge, this is the first study on grounding textual representations in pause signals for improved understanding in voice assistants.

%% file: 07-acknowledgements.tex
\section*{Acknowledgements}
\label{section:acknowledgements}

The authors would like to thank Barry Theobald, Murat Akbacak, Lin Li, Sachin Agarwal, Anusha Kamath, John Bridle, Aaksha Meghawat, Melvyn Hunt, and Ahmed Hussen Abdelaziz for their help and feedback.